\documentclass[10pt,twocolumn,letterpaper]{article}

\usepackage{iccv}
\usepackage{times}
\usepackage{epsfig}
\usepackage{graphicx}
\usepackage{amsmath}
\usepackage{amssymb}
\usepackage{caption}
\usepackage{subcaption}
\usepackage{multirow}
\usepackage{paralist}
\usepackage[pagebackref=true,breaklinks=true,letterpaper=true,colorlinks,bookmarks=false]{hyperref}

\iccvfinalcopy

\ificcvfinal\fi
\begin{document}
\title{Efficient Large Scale Video Classification}
\author{Balakrishnan Varadarajan \\
  Google, Inc. \\
  {\tt\small balakrishnanv@google.com} \\
  \and
  George Toderici \\
  Google, Inc. \\
  {\tt\small gtoderici@google.com} \\
  \and
  Sudheendra Vijayanarasimhan \\
  Google, Inc. \\
  {\tt\small svnaras@google.com} \\
  \and
  Apostol Natsev \\
  Google, Inc. \\
  {\tt\small natsev@google.com} \\
}
\maketitle
\begin{abstract}

Video classification has advanced tremendously over the recent years. A large part of the improvements in video classification had to do with the work done by the image classification community and the use of deep convolutional networks (CNNs) which produce competitive results with hand-crafted motion features. These networks were adapted to use video frames in various ways and have yielded state of the art classification results. We present two methods that build on this work, and scale it up to work with millions of videos and hundreds of thousands of classes while maintaining a low computational cost. In the context of large scale video processing, training CNNs on video frames is extremely time consuming, due to the large number of frames involved. We propose to avoid this problem by training CNNs on either YouTube thumbnails or Flickr images, and then using these networks' outputs as features for other higher level classifiers. We discuss the challenges of achieving this and propose two models for frame-level and video-level classification. The first is a highly efficient mixture of experts while the latter is based on long short term memory neural networks. We present results on the Sports-1M video dataset (1 million videos, 487 classes) and on a new dataset which has 12 million videos and 150,000 labels.

\end{abstract}

 \vspace{-0.5em}
\section{Introduction}
 \vspace{-0.5em}
\label{sec:intro}

Video classification is the task of producing a label that is relevant to the
video given its frames. A good video level classifier is a one that not only
provides accurate frame labels, but also best describes the entire video given
the features and the annotations of the various frames in the video. For
example, a video might contain a tree in some frame, but the label that is
central to the video might be something else (e.g., ``hiking''). The
granularity of the labels that are needed to describe the frames and the video
depends on the task. Typical tasks include assigning one or more global labels
to the video, and assigning one or more labels for each frame inside the video.
In this paper we deal with a truly large scale dataset of videos that best
represents videos in the wild. Much of the advancements in object recognition
and scene understanding comes from convolutional neural networks
~\cite{liris2011,ji2013,karpathy2014large,simonyan2014two}. The key factors
that enabled such large scale success with neural networks were improvements in
distributed training, advancements in optimization techniques and architectural
improvements\cite{szegedy14going}. While the best published
results~\cite{lan2014beyond} on academic benchmarks such as UCF-101 use motion
features such as IDTF~\cite{wang11}, we will not make use of them
in this work due to their high computational cost.

\begin{figure}[tbp]
  \includegraphics[width=0.5\textwidth]{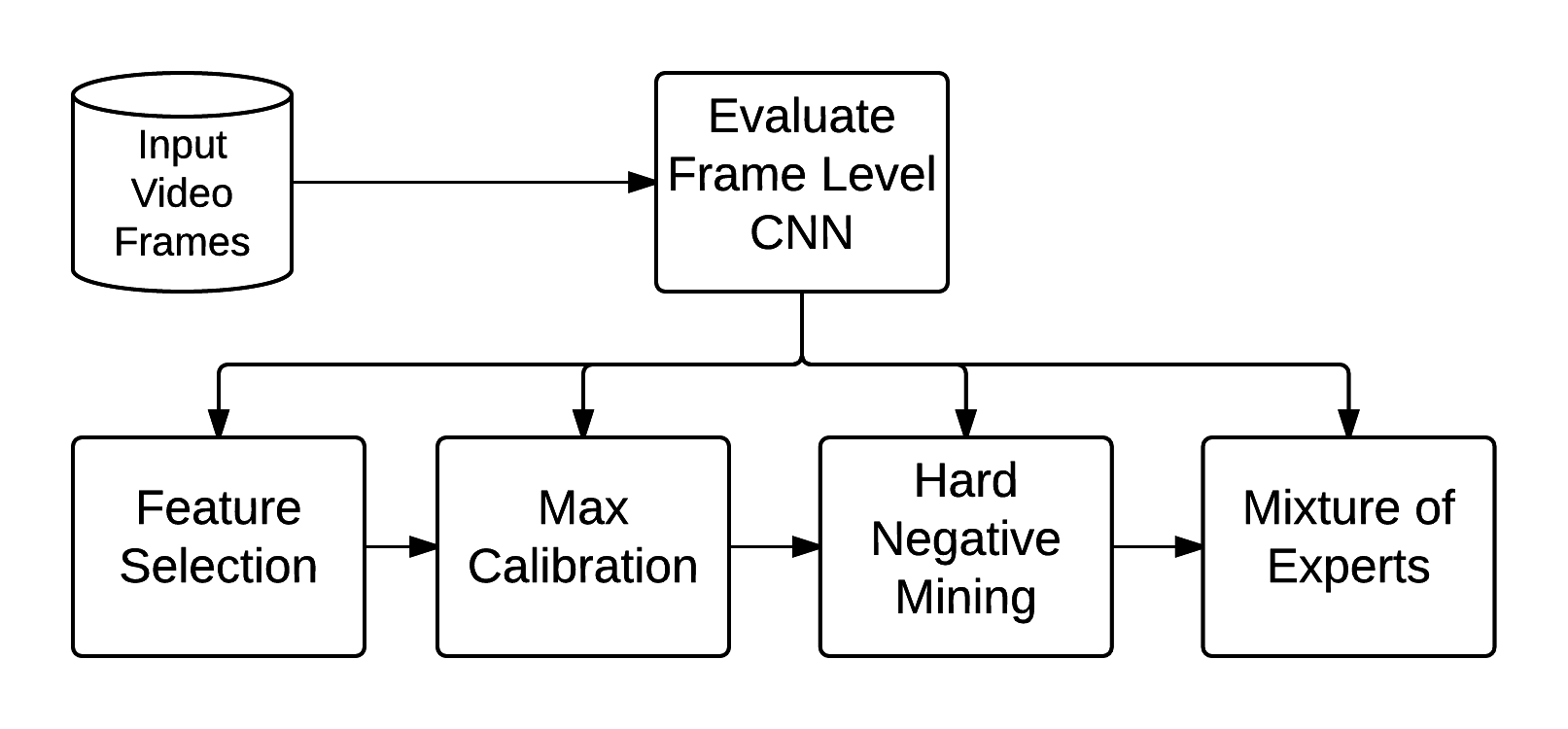}
  \caption{Overview of the MiCRObE training pipeline.}
  \label{fig:microbe}
\end{figure}

Training neural networks on video is a very challenging task due to the large
amount of data involved. Typical approaches take an image-based network, and
train it on all the frames from all videos in the training dataset. We created
a benchmark dataset on which this would simply be infeasible. In our dataset we
have 12 million videos. Assuming a sampling rate of 1 frame per second, this
would yield 2.88 billion frames. Training an image-based on such a large number
of images would simply take too long with current generation hardware. Another
challenge which we aim to address is how to handle a very large number of
labels. In our dataset we have 150,000 labels.

We approach the problem of training using such a video corpus using two key
ideas: 1) we use CNNs that were trained using video thumbnails or Flickr images
as base features; and 2) the scale is large enough that only distributed algorithms
may be used. Assuming a base image-based CNN classifier of 150,000 classes, and that
on average 100 of these classes trigger per frame, an average video in our
dataset would be represents using 24,000 features. In a na\"{i}ve linear classifier
this may require up to 3.6 billion weight updates. Assuming a single pass over
the data, in the worst case it would generate $43 \times 10^{15}$ updates.

The main contribution of this paper is describing two orthogonal methods which
can be used to learn efficiently on such a dataset. The first method consists
of a cascade of steps. We propose to use an initial relatively weak classifier
to quickly learn feature-class mappings while pruning as many of these
correlations as possible. This classifier is then used for hard negative mining
for a second order classifier which then is improved iteratively. The second method
employs an optimized neural network architecture using long short-term memory
(LSTM) neurons~\cite{hochreiter97long} and hierarchical
softmax~\cite{morin2005hierarchical}, while using a distributed training
architecture~\cite{dean2012large}.

We present two methods for both frame-level and video-level classification. The
first, named MiCRObE (Max Calibration mixtuRe Of Experts, see Figure~\ref{fig:microbe}) is described in
Section~\ref{sec:microbe}, while the second method which we abbreviate as LSTM
is described in Section~\ref{sec:lstm}.

\vspace{-0.5em}
\section{Related Work}
\vspace{-0.5em}
\label{sec:related}
Our work is targeted at large scale video classification. In terms of public benchmarks, the largest publicly available benchmark is Sports-1M~\cite{karpathy2014large}, which contains over one million videos, and 487 labels. The best performing classification method on the Sports-1M benchmark has been using a frame-level convolutional deep neural network, with either max-pooling or LSTM on top~\cite{ng2015beyond}. Using the same benchmark, Karpathy~\etal~\cite{karpathy2014large}, and Tran~\etal~\cite{tran14corr} propose using a convolutional deep network for making frame and video-level predictions, while Karpathy~\etal~\cite{karpathy2014large} also present results on using hand-crafted features and deep networks. The inputs to these networks is raw pixels, and the networks are trained through convolutional part, resulting in a very long training time. Other large scale video classification methods~\cite{aradhye09,toderici10,yang2011discriminative} used hand-crafted features and per-class AdaBoost classifiers, but were only able to use a fraction of the videos to train the per-class models. Unlike previous work, our goal is to provide fast training times with models capable of frame-level and video-level prediction, while allowing for a much larger number of labels and videos to be used for training.

Many of the best performing models in machine learning problems such as image classification, pattern recognition and machine translation come by fusion of multiple classifiers \cite{sebastiano_fusion,saenko_fusion}.  Given scores $p_l^{j}$ for $j=1,\ldots,M$ from each of the $M$ sources for a label $l$, the traditional fusion problem is a function $\hat{p}_l = f(p_l^{1:M})$ that maps these probabilities to a single probability value. This problem is well studied and one of the most popular techniques is Bayes fusion which has been successfully applied to vision problems~\cite{Kittler98e,manduchi_cvpr}. Voting based fusion techniques like majority and sum based  are extremely popular mostly because of they are simple and non parametric. The current best result on image net is based on a simple ensemble average of six different classifiers that output ImageNet labels~\cite{sergey_ensemble}.

The fundamental assumption in these settings is that each of the $M$ sources need to speak the same vocabulary as that of the target. What if the underlying sources do not speak the same vocabulary, yet output semantically meaningful units? For example, the underlying classifier only detected \emph{canyon}, \emph{river} and \emph{rafting}. Can we learn to infer the probability of the target label being \emph{Grand Canyon} from these detections? Another extreme is to have the underlying classifier so fine-grained that it has (for example) the label \emph{African elephant}, but does not have the label \emph{elephant}. If the label \emph{elephant} is present in the target vocabulary, can we learn to infer the relation \emph{African elephant} $\Rightarrow$ \emph{elephant} organically from the data? One approach is to treat the underlying classification outputs as \emph{features} and train the classifiers for each label based on these features. This idea has been proposed in the context of scene classification~\cite{li2010object}. This approach can quickly run into the curse of dimensionality, especially if the underlying feature space is huge (which is indeed the case for our problem).

%


\vspace{-0.5em}
\section{MiCRObE}
\label{sec:microbe}
\vspace{-0.5em}
\subsection{Feature Calibration}
\vspace{-0.5em}

The process of calibration is to learn a one dimensional model that computes the probability of each label $e$ given a single feature $f$. Here, the sparse features can be semantic units that may or may not speak in the same vocabulary as the target labels. In addition to providing a simple $\operatorname{max-calibration}$ based label classifier, the calibration process also helps in feature selection that can be used to significantly speed  up the training of classifiers like SVM or logistic regression. The feature selection process yields:
(a) Automatic synonym expansion using visual similarity: As an example, we allow the sparse feature named \emph{Canyon} from one of our base models to predict an entity \emph{Grand Canyon} which is not in the set of input sparse features. Similarly \emph{Clock Tower} will be able to predict \emph{Big Ben}.
(b) Automatic expansion to related terms based on visual co-occurrence: For example, we will get the feature \emph{water} for the label \emph{boat} which can be used as a supporting evidence for the boat classifier.\newline
In other words, ``Canyon'', ``Clock Tower'', ``cooking'', ``water" are features but ``Grand Canyon'', ``boat" and ``Big Ben'' are labels. Formally put, the number of input features is a sparse 150,000 dimensional vector which is a combination of predictions from various classifiers. The output is a target label set of  labels. The calibration model is a function $p_{e|f}(x)$ that is defined over pairs of label ($e$) and feature ($f$) that is learned according to an isotonic regression. We use a modified version of the Platt's scaling~\cite{Platt99probabilisticoutputs} to model this probability:
\begin{equation}
p_{e|f}(x) = \alpha \left(\sigma(\beta x + \gamma) - \sigma(\gamma) \right)
\end{equation}
where $\sigma(x)=\frac{1}{1+\exp(-x)}$ is the sigmoid function and $\alpha, \beta, \gamma$ are functions of $e$ and $f$. We enforce $\alpha,\beta \geq 0$ so that the function $p_{e|f}(x)$ monotonically increases with $x$ (the feature value). Furthermore, since $p_{e|f}(x)$ is a probability, we need to enforce that $p_{e|f}(x_{\operatorname{max}}) \leq 1$ where $x_{\operatorname{max}}$ is the maximum feature value from the training data. The scale $\alpha$ allows the estimated probability to plateau to a value less than 1.0 (a property that cannot be enforced in normal Platt's scaling). For example, one of the input sparse feature is  the detection ``Canyon'' from a base image classifier. There are at least a dozen canyon's in the entire world (including Grand Canyon). It is reasonable for the probability of grand canyon to have a value less than 1.0 even if the input sparse feature ``Canyon'' fired with the highest confidence from an extremely precise base classifier. Furthermore, the offset term enforces that $p_{e|f}(0)=0$, which helps when dealing with sparse features. Thus, we only capture positively correlating features for a label $e$. Fitting of $p_{e|f}(x)$ can be either done by minimizing the squared error or the log-loss over all instances (video-frames in our case) where $x_f> 0$. We used squared loss in our implementations as we found it to be more robust near the boundaries, especially given that $p_{e|f}(x)$ is enforced to zero. For each instance where $x_f > 0$ we also have a ground-truth value associated with label $e$ as $g_e$. Given training examples $(w_t, x_t, g_t)_{t \in T}$ where $w_t$ is the weight of the example\footnote{To speed up the implementation, we quantize the feature values in buckets of size $10^{-4}$ and the weight $w$ is the total number of examples that fell in that bucket and $g$ is the mean ground-truth value in that bucket.}, $x_t$ is the feature value and $g_t$ is the ground-truth, we estimate $\alpha, \beta, \gamma$ by solving the following regularized least squares
\begin{equation}
(\hat{\alpha}, \hat{\beta}, \hat{\gamma}) = \operatorname{argmin} \sum_{t \in T} w_t (p_{e|f}(x_t) - g_t)^2 + \lambda (\alpha^2 + \beta^2 + \gamma^2)
\end{equation}
subject to $\alpha \geq 0$ and $\beta \geq 0$. $\lambda$  is tuned on a held out set to minimize the held out squared loss. We estimate  9 billion triples of $(\alpha, \beta, \gamma)$ and only retain the ones where the estimated $\alpha> 0$.
 Since the problem has only 3 variables, we can compute the exact derivative and Hessian w.r.t. $\alpha, \beta, \gamma$ at each point and do a Newton update. The various $(e,f)$ pairs are processed in parallel. Once the function $p_{e|f}(x)$ is learned, we choose up to the top K features sorted according to $p_{e|f}(x_{max})$ (the maximum probability of the label given that the feature). The outcome is a set  $F_e$ of positively correlated features for each label $e$.

 \vspace{-0.5em}
\subsection{Max Calibration Model}
 \vspace{-0.5em}
Once the calibrations $p_{e|f}(x)$ are learned for (label, feature) pairs, the $\operatorname{max-calibration}$ model is an optimistic predictor of the probability of each entity $e$ given the set of all features that fired in the frame $x$ as
\begin{equation}
p_e(x) = \max_{f} p_{e|f}(x_f)
\label{eqn:maxcal}
\end{equation}
Note that the max calibration model works best when the input features are sparse outputs that have some semantic meaning. Despite the simplicity and robustness of the $\operatorname{max-calibration}$ model, there are several drawbacks that may limit it from yielding the best performance:\newline
(a) The max calibration model uses noisy ground truth data (assumes all frames in the video are associated with the label). At the very least, we need to correct this by learning another model that uses a cleaner ground truth. \newline
(b) Furthermore, the non-linear operation of doing a max on all the probabilities may result in overly optimistic predictions for labels which have a lot of features $F_e$. Hence the output will no longer be well calibrated (unless we learn another calibrator on top of the max-calibrated score). \newline
(c) Each feature is treated independently, hence the $\operatorname{max-calibration}$ model cannot capture the correlations between the various features. Max calibration model can only deal with sparse features. For example, we cannot use continuous valued features like the output of an intermediate layer from a deep network.

As a result, we will use the max-calibration model as a bootstrapping mechanism for training a second order model.

\textbf{Hard Negative Mining:} The $\operatorname{max-calibration}$ model provides calibrated probabilities for all labels in our vocabulary. It is a simple model and is extremely efficient to compute. Hence, we will exploit this property to mine good positives and hard-negatives (i.e., the ones that score high according to the $\operatorname{max-calibration}$ model). The mining process for an entity $e$ can be described formally as sorting (from highest to lowest) all the training examples (video frames in our case) according to the $\operatorname{max-calibration}$ score of $e$ and retaining the top $M$ examples. We chose $M$ such that it captures more than 95\% of the positives. Since the number of training examples is huge (e.g., 3.6 billion frames, in our case), we do this approximately using map-reduce where, in each of the W workers, we collect the top $k$ examples and choose the top M examples from the resulting $k W$ examples. Although this approach is approximate, if $kW$ is chosen to be sufficiently larger than $M$, we can guarantee that we can recover almost all of the top $M$ examples. The expected number of the true top $M$ examples that will be recovered by choosing the top $M$ examples from this $kW$ sized set is given as
\begin{equation}
  \small
E(k, W, M) = k + \sum_{i=k+1}^{M} \left(1 - \frac{1}{W} \right)^{i-1} \sum_{j=0}^{k-1} \binom{i-1}{j} (W-1)^{-j}
\end{equation}
For example if, $M=80000$ examples, $W=4000$ workers and $k = 40$ examples/worker, this evaluates to $79999.8126$. In general, setting $kW=2M$ yields a good guarantee. In the next section, we show how to get the top $k$ examples from each worker efficiently.

 \vspace{-0.5em}
\subsection{Choosing Top-$k$ Examples per Worker}
 \vspace{-0.5em}
The brute force approach to achieve this is to compute the max calibration score using (\ref{eqn:maxcal}) for each label $e$ given the features $\mathbf{x}$ for all examples that belong to the worker $w$ and insert ${p_e(\mathbf{x}), e, \mathbf{x}}$ into a $k$-sized priority queue (which is keyed by the max calibrated probability $p_e(\mathbf{x})$) for the label $e$. Unfortunately, this can be very time consuming, especially when assigning millions of examples per worker. In this section, we propose an approach that makes this mining extremely efficient and is particularly tuned towards the $\operatorname{max-calibration}$ model. The idea is to only score labels which are guaranteed to enter the priority queue. As a result of this, computing $p_{e|f}(x_f)$ becomes less and less frequent as more examples are processed in the worker and the priority queue for $e$ continues to get updated.

From the calibration step, we have a set of shortlisted features $F_e$ for each label $e$. If we invert this list, we get a set of shortlisted labels $E_f$ for each feature $f$. In each worker $w$, we also maintain a priority queue $\mathbf{Q}(e, w)$ for each label $e$ that stores up to the top-$k$ examples (according to the $\operatorname{max-calibration}$ score).

In each worker $w$, for each feature $f$, we store an inverse lookup to labels $E_f(w)$ which is initially $E_f$.  In addition, we also store a minimum feature firing threshold $\tau_{f,e}$ such that only if $(x_f \geq \tau_{f,e})$ for some $f$, we will insert $e$ into the priority queue. Initially $\tau_{f,e} = 0$, which implies that every label $e \in E_f$ for all $f$ such that $x_f > 0$ will be scored.

Let the minimum $\operatorname{max-calibration}$ value in the priority queue be $Q_{min}(e,w)$. This is zero if the size of the priority queue is less than $k$, otherwise (when the size is equal to $k$) it is the smallest element in the queue.

For each training example, let $\mathbf{x}$ be the sparse feature vector and $g$ be the corresponding ground-truth.

Let $p_e(\mathbf{x})$  be the score of $e$ according to the $\operatorname{max-calibration}$ model for this instance. Instead of computing $p_e(\mathbf{x})$ explicitly for \emph{all} labels using the $\operatorname{max-calibration}$ model, we only compute it for a subset of the labels which are guaranteed to enter the priority queue $\mathbf{Q}(e,w)$ as follows: $p_e(\mathbf{x})$ is initially zero for all $e$ (an empty map). For each feature $f  : x_f > 0$ and  for each label $e \in E_f(w)$, if $x_f < \tau_{f,e}$ and  $p_{e|f}(x_f) < Q_{min}(e)$, we update $\tau_{f,e}$ to $x_f$. In addition if $p_{e|f}(x_f) < Q_{min}(e)$, we remove $e$ from $E_f(w)$ (so we have fewer labels in the inverse lookup for $f$). On the other hand,  if $x_f \geq \tau_{f,e}$ and $p_{e|f}(x_f) \geq Q_{min}(e)$, we update $p_e(\mathbf{x})$ to $\max(p_e(\mathbf{x}), p_{e|f}(x_f))$. For each $e$ where $p_e(\mathbf{x}) > 0$, insert $\{p_e(\mathbf{x}), e, \mathbf{x}, g\}$  to the priority queue $Q(e, w)$.

These $M$ examples become the training data for another second order classifier. Since the second order model is trained only on these $M$ examples, it is important to retain the distribution at inference time. The second order model may not do well with an odd-man-out data point (that is not in the distribution of these $M$ examples) is seen. Hence at inference time, we put a threshold on the lowest $\operatorname{max-calibration}$ score of any positive example seen in the $M$ training examples.

\textbf{Popular Labels:} For popular YouTube labels like \emph{Minecraft}, $M$ needs to be sufficiently high to capture a significant fraction of the positives. For example, \emph{Minecraft} occurs in $3\%$ of YouTube videos in our set. On a dataset of $10$ million videos with frames sampled at 1fps and each video having an average length of 4 minutes, we have $72$ million frames that correspond to \emph{Minecraft}. In this case $M$ needs to be much higher than $72$ million and this is not feasible to fit in a single machine.\footnote{The second order classifier is trained in parallel across different workers, but we train each label in a single machine. }  When considering each example for such labels to be added to the top-$k$ list in each worker, we do a random sampling with a probability $p$ which is proportional to $\frac{M}{positives(label)}$ [i.e., the step ``insert $\{p_e(\mathbf{x}), e, \mathbf{x}\}$ to the priority queue'' is done with this probability].

 \vspace{-0.5em}
\subsection{Training the Second Order Model}
 \vspace{-0.5em}
Given the top $M$ examples of positives and negatives obtained from the hard negative mining using the first order $\operatorname{max-calibration}$ model,  the second order model learns to discriminate the good positives and hard-negatives in this set. At inference time, for each example $X$, we check if the $\operatorname{max-calibration}$ score is at least as much as the $\operatorname{max-calibration}$ score of any positive example in this training set. Note that checking if the $\operatorname{max-calibration}$ is at least $\tau_e$ is equivalent to checking if at least one of the feature values passes a certain threshold. Formally put

\begin{equation}
\max_{f} p_{e|f}(x_f) \geq \tau_e \equiv \bigcup_{f} I(x_f \geq p_{e|f}^{-1}(\tau_e))
\end{equation}

\noindent $p_{e|f}(x_f)$ is monotonically increasing and hence its inverse is uniquely defined. At the inference time, we check if at least one feature exceeds the certain threshold $\eta_{e|f} = p_{e|f}^{-1}(\tau_e)$ which is pre-computed during initialization. If the $\operatorname{max-calibration}$ score exceeds this threshold, we apply the second order model  $Q_e(\mathbf{x})$ to compute the final score of the label $e$ for the example $\mathbf{x}$. If the max calibration score does not exceed this threshold, the final score $P_e(\mathbf{x})$ of the label $e$ is set to zero. This is essentially a 2-stage cascade model, where a cheap max calibration model is used as an initial filter, followed by a more accurate and more expensive second-order model. We used logistic regression and mixture of experts as candidates for this second order model.

 \vspace{-0.5em}
\subsection{Mixture of Experts}
 \vspace{-0.5em}
Recall that we train a binary classifier for each label $e$. $y=1$ denotes the existence of $e$ in the features $\mathbf{x}$. Mixture of experts (MoE) was first proposed by Jacobs and Jordan \cite{Jordan94hierarchicalmixtures}. In an MoE model, each individual component models a different binary probability distribution. The probability according to the mixture of $H$ experts is given as
\begin{equation}
p(y = 1 | \mathbf{x}) = \sum_{h} p(h | \mathbf{x}) p(y = 1 | \mathbf{x}, h)
\end{equation}
where the conditional probability of the hidden state given the features is a soft-max over $H+1$ states $p(h | \mathbf{x}) = \frac{\exp(\mathbf{w}_h^{T} \mathbf{x})}{1 + \sum_{h'} \exp(\mathbf{w}_{h'}^{T} \mathbf{x})}$. The last $(H+1)^{th}$ state is a dummy state that always results in the non-existence of the entity. When $H=1$, it is a product of two logistics and hence is more general than a single logistic regression. The conditional probability of $y=1$ given the hidden state and the features is a logistic regression given as $p(y = 1 | \mathbf{x}, h) = \sigma(\mathbf{u}_h^{T} \mathbf{x})$.  The parameters to be estimated are the softmax gating weights $\mathbf{w}_h$ for each hidden state and the expert logistic weights $\mathbf{u}_h$. For the sake of brevity, we will denote $p_{y|x} = p(y=1 | \mathbf{x})$, $p_{h|\mathbf{x}} = p(h | \mathbf{x})$ and $p_{h} = p(y=1 | \mathbf{x},h)$ Given a set of training data ${(\mathbf{x}_i, g_i)}_{i=1\ldots N}$ for each label where $\mathbf{x}_i$ is the feature vector and $g_i$ is the corresponding boolean ground-truth, we optimize the regularized loss of the data which is given by
\begin{equation}
\sum_{i=1}^{N}  w_i \mathcal{L} \left[p_{y|\mathbf{x}_i}, g_t\right] +\lambda_{2} \left(\Vert \mathbf{w} \Vert_{2}^2 + \Vert \mathbf{u} \Vert_{2}^2 \right)
\label{eqn:loss}
\end{equation}
where the loss function $\mathcal{L}(p,g)$ is the log-loss given as
\begin{equation}
\mathcal{L}(p,g) = -g \log p - (1-g) \log (1-p)
\label{eqn:logloss}
\end{equation}
and $w_i$ is the weight for the $i^{th}$ example.

\textbf{Optimization:}
Note that we could directly write the derivative of $\mathcal{L} \left[p_{y|\mathbf{x}}, g\right]$ with respect to the softmax weight $\mathbf{w}_h$ and the logistic weight $\mathbf{u}_h$ as
\begin{eqnarray*}
\frac{\partial \mathcal{L} \left[p_{y|\mathbf{x}}, g\right]}{\partial \mathbf{w}_h} &=& \mathbf{x} \frac{p_{h|\mathbf{x}} \left(p_{y|h,\mathbf{x}} - p_{y|\mathbf{x}}\right) \left(p_{y|\mathbf{x}}-g\right)}{p_{y|\mathbf{x}}(1-p_{y|\mathbf{x}})} \\
\frac{\partial \mathcal{L} \left[p_{y|\mathbf{x}}, g\right]}{\partial \mathbf{u}_h} &=& \mathbf{x} \frac{p_{h|\mathbf{x}} p_{y|h,\mathbf{x}} (1- p_{y|h,\mathbf{x}})\left(p_{y|\mathbf{x}}-g\right)}{p_{y|\mathbf{x}}(1-p_{y|\mathbf{x}})}
\end{eqnarray*}
Our implementation uses the \textit{ceres} library~\cite{ceres-solver} to solve the minimization in (\ref{eqn:loss}) to obtain the weights $(\mathbf{w}_h, \mathbf{u}_h)$ using the Broyden Fletcher Goldfarb Shanno algorithm (LBFGS). We also implemented an EM variant where the collected statistics are used to re-estimate the softmax and the logistic weights (both of which are convex problems). However, in practice, we found that LBFGS converges much faster than EM and also produces better objective in most cases. For all our experiments, we report accuracy numbers using the LBFGS optimization.

\textbf{Initialization:} When $H$ (the number of mixtures) is greater than one, we select $H$ positive examples according to the non-deterministic KMeans$++$ sampling strategy~\cite{arthur2007k}. The features of these positive examples become the gating weights (the offset term is set to zero). The expert weights are all initialized to zero. We then run LBFGS until the relative change in the objective function ceases to exceed $10^{-6}$. When $H=1$, we initialize the expert weights to the weights obtained by solving a logistic regression, while the gating weights are all set to zero. Such an initialization ensures that the likelihood of the trained MoE model is at least as much as the one obtained from the logistic regression. In our experiments, we also found consistent improvements by using MoE with 1 mixture compared to a logistic regression and small improvements by training a MoE with (up to) $5$ mixtures compared to a single mixture. Furthermore, for multiple mixtures, we run several random restarts and pick the one that has the best objective.

\textbf{Hyperparameter Selection:} In order to determine the best $L_2$ weight $\lambda$ on $\mathbf{w}_h$ and $\mathbf{u}_h$, we split the training data into two equal video dis-joint sets and grid search over $\lambda$ and train a \emph{logistic regression} with an $L_2$ weight of $\lambda$. For our experiments, we start $\lambda=10^{-2}$ and increase it by a factor of $2$ in each step. Once we find that the holdout loss is starting to increase, we stop the search.

\textbf{Training times:}The total training time (from calibration to training the MoE model) for training the frame-level model takes less than 8 hours by using all features on the 10.8 million training videos by distributing the load across 4000 machines. When the number of mixtures is greater than one, the majority of the training time is spent doing the random restarts and hyper-parameter sweep. The corresponding training time for the video level model is anywhere between twelve to sixteen hours on the 12 million set. Training the same models on the sports videos takes less than an hour. Inference takes $\leq1$s per 4-minute video.

\vspace{-1em}
\section{Video and Frame-Level Prediction LSTM}
\vspace{-0.5em}
\label{sec:lstm}

We also tackle the task of frame-level and video-level prediction using recurrent neural networks. In this section we describe our approach.

A recurrent network operates over a temporally ordered set of inputs $\boldsymbol{x} = \left\{x_1, \dotsc, x_T\right\}$. $x_t$ corresponds to the features at time step $t$. For each time step $t$ the network computes a hidden state $h_t$ which depends on $h_{t-1}$ and the current input, and bias terms. The output $y_t$ is computed as a function of the hidden state at the current time step:

\begin{align}
  h_t &=  \mathcal{H}(W_{x}x_t + W_hh_{t-1} + b_h) \\
  y_t &= W_oh_t + b_o
\end{align}

\noindent where $W$ denote the weight matrices. $W_x$ denotes the weight matrix corresponding to the input features, $W_h$ denotes the weights by which the previous hidden state is multiplied, and $W_o$ denotes the weight matrix that is used to compute the output. $b_h$ and $b_o$ denote the hidden and output bias. $\mathcal{H}$ is the hidden state activation function, and is typically chosen to be either the sigmoid or the $\tanh$ function.

This type of formulations suffers from the vanishing gradient
problem~\cite{bengio94learning}.  Long Short-Term Memory neurons have been
proposed by Schmidhuber~\cite{hochreiter97long} as type of neuron which does
not suffer from this. Thus, LSTM networks can learn longer term dependencies
between inputs, which is why we chose to use them for our purposes.

The output of the hidden layer $h_t$ for LSTM is computed as follows:

\begin{align}
i_t &= \sigma (W_{xi}x_t + W_{hi}h_{t-1} + W_{ci}c_{t-1} + b_i) \label{eqi}\\
f_t &= \sigma (W_{xf}x_t + W_{hf}h_{t-1} + W_{cf}c_{t-1} + b_f) \label{eqf}\\
c_t &= f_tc_{t-1} + i_t\ \tanh(W_{xc}x_t + W_{hc}h_{t-1} + b_c) \label{eqc}\\
o_t &= \sigma (W_{xo}x_t + W_{ho}h_{t-1} + W_{co}c_t + b_o) \label{eqo}\\
h_t &= o_t\ \tanh(c_t) \label{eqh}
\end{align}

\noindent where $\sigma$ is the sigmoid function. The main difference between
this formulation and the RNN is that the $i_t$ decides whether to use the input
to update the state, $f_t$ decides whether to forget the state, and $o_t$
decides whether to output.  In some sense, this formulation introduces data
control flow driven by the state and input of the network.

For the first time step, we set $c_{0} = \mathbf{0}$ and $h_{0} = \mathbf{0}$. However,
the initial states could also be represented by using a learned bias term.

For the purposes of both video and frame-level classification, we consider a
neural network which has frame-level classifications as inputs. These scores
are be represented in a sparse vector $\boldsymbol{s_t} = \left\{s_t^i | \forall s_t^i > 0, s_t^i \in S_t\right\}$, where $S_t$ is a vector containing the scores for all classes at time $t$. The first layer of the network at time $t$ computes its output as $x_t = \sum_i s_t^i w^i + b$, where $b$ is the bias term. This formulation of $s_t$, significantly reduces the amount of computation needed for both the forward and backward pass because it only considers those elements of $S_t$ which have values greater than zero. For our networks, the number of non-zero elements per frame is less than 1\% of the total possible. In or experiments, each class is represented internally (as $w^i$) with 512 weights.

On top of this layer, we stack 5 LSTM layers with 512 units each~\cite{graves2013speech}. We unroll the LSTM layers for 30 time steps, which is equivalent to using 30 seconds' worth of video at a time for training. Each of the top LSTM layers is further connected to a hierarchical softmax layer~\cite{morin2005hierarchical}. In this layer, we use a splitting factor of 10, and a random tree to approximate the hierarchy.

Similarly to~\cite{ng2015beyond} we use a linearly increasing weight for each time step, starting with $1/30$ for the first frame, and assigning a weight of $1$ to the last frame. This allows the model to not have to be penalized heavily when trying to make a prediction with few frames. We also investigated using an uniform weight for each frame and max-pooling over the LSTM layer, but in our video-level metrics, these methods proved inferior to the linear weighting scheme.

In our dataset, videos have an average of 240 seconds. Therefore, when using the 30-frame unrolled LSTM model, it is not clear what to do in order to obtain a video-level prediction. In order to process the entire video, we split it into 30-second chunks. Starting with the first chunk of the video, we predict at every frame, and save the state at the end of the sequence. When processing subsequent chunks, we feed the previously saved state back into the LSTM. At the end of the video we have as many sets of predictions as we have frames. We investigated using max-pooling and average pooling over the predictions, and as an alternative, taking the prediction at the last frame.

For every video, our LSTM model produces a 512-dimensional representation. This is the state of the top-most LSTM layer at the last frame in the video. This also allows other classifiers to be trained using this representation.

The training was done using distributed stochastic gradient descent~\cite{dean2012large} using 20 model replicas. We used a learning rate of 0.3, and employed the AdaGrad update rule~\cite{duchi2011adaptive}. The training took less than 5 days before convergence. Inference takes $\leq1.4$ seconds for the average 4-minute video.

\vspace{-2em}
\section{Experimental Setup}
\vspace{-0.5em}
\textbf{Datasets:} We created a new dataset of \textbf{12 million YouTube videos} spanning about $150,000$ visual labels from Freebase~\cite{bollacker2008freebase}. We selected these 12 million videos such that each of them have a view count of at least $10,000$. The $150,000$ labels were selected by removing music topics such as songs, albums and people (to remain within the visual domain and not having to concentrate on face recognition). YouTube provides the labels of the videos which obtained by running a Freebase-based annotator \cite{simonet-wole-13} on title, description and other metadata sources. We retrieved the videos belonging to each label by using the YouTube Topics API~\cite{youtube_topics}. This annotation is fairly reliable  for high view videos where the weighted precision is over $95\%$ based on human evaluation. Many of the labels are however extremely fine-grained, making them visually very similar or even indistinguishable. Some examples are \textit{Super Mario 1}, \textit{Super Mario 2}, \textit{FIFA World Cup 2014}, \textit{FIFA World Cup 2015}, \textit{african elephant}, \textit{asian elephant}, etc. These annotations are only available at the video-level. Another dataset that we used for is  \textbf{Sports-1M dataset} \cite{karpathy2014large} that consists of roughly 1.2 million YouTube sports videos annotated with 487 classes. We will evaluate our models both at the video level and the frame level.

\textbf{Features:} We extract two sets of sparse features from each frame (sampled at 1 fps) for the videos in our training and test set. One set of features are the prediction outputs of an \textit{Inception}-derived deep neural network \cite{szegedy14going} trained on \emph{YouTube thumbnail images}. This model by itself performs much worse on our training set, because YouTube thumbnails are noisy (tend to be visually more attractive than describing the concept in the video) and is only a single frame from the entire YouTube video. The number of unique sparse features firing from this model on our 10 million training set is about $110,000$. In our experiments section, we will abbreviate these features as \textbf{TM} which stands for \emph{thumbnail model}. Another set of features are the predictions of a deep neural network with a similar architecture as the \textbf{TM} model, but is largely trained on \emph{Flickr data}. The target labels are again from the metadata of the Flickr photos and are similar spirit to image net labels~\cite{krizhevsky2012imagenet}. Moreover, they are much less fine-grained than the YouTube labels. The vocabulary size of these labels is about $17,000$. For example, the label \emph{Grand Canyon} won't be present. Instead the label \emph{Canyon} will be present.  We will abbreviate these features as \textbf{IM} that stands for \emph{Image models}.

For both models we process the images by first resizing them to $256 \times 256$ pixels, then randomly sampling a $220 \times 220$ region and randomly flipping the image horizontally with $50\%$ probability when training. Similarly to the LSTM model, the training was performed on a cluster using Downpour Stochastic Gradient Descent~\cite{dean12} with a learning rate of $10^{-3}$ in conjunction with a momentum of $0.9$ and weight decay of $0.0005$.
\vspace{-0.5em}

\subsection{Training and Evaluation}
\vspace{-0.5em}
\begin{table*}
\begin{center}
\begin{tabular}{|c|c|c|c|c|c|c|c|c|c|}
\hline
Features & Dataset & \textbf{Self} &  \textbf{MaxCal}& \shortstack{\textbf{Logit} \\ \small{Random Negs.}} & \shortstack{\textbf{Logit} \\ \small{Hard Negs.}} & \shortstack{\textbf{MiCRObE} (1 mix) \\ \small{Hard Negs}.} & \shortstack{\textbf{MiCRObE} (5 mix) \\ \small{Hard Negs}.} \\
\hline
\textbf{IM} & \multirow{3}{*}{YT-12M}  & 4.0\% & 20.0\% & 27.0\% & 29.2\% & 31.3\% & 32.4\% \\
\cline{1-1}\cline{3-8}
\textbf{TM} & ~ & 19.0\% &  28.0\% & 31.0\% & 39.8\% &  40.6\% & 41.0\% \\
\cline{1-1}\cline{3-8}
\textbf{IM}$+$\textbf{TM} &  ~ & 7.0\% &33.0\% & 40.6\% & 42.5\% & 43.9\% & 43.8\% \\
\hline
\textbf{IM} & \multirow{3}{*}{Sports-1M} &  0.9\% & ~ & 25.6\% & 35.0\% & 39.3\% & 39.8\% \\
\cline{1-1}\cline{3-8}
\textbf{TM} & ~ & 1.2\% & ~ & 33.9\% & 45.7\% &  46.8\% & 49.9\% \\
\cline{1-1}\cline{3-8}
\textbf{IM}$+$\textbf{TM} & ~ & 1.5\% & 39.0\% & 41.0\% & 47.8\% & 49.8\% & 50.2\% \\
\hline
\end{tabular}
\end{center}
\caption{Frame level model evaluation against the video-level ground truth. The values in the table represent hit@1.}
\label{tab:afp}
\vspace{-1em}
\end{table*}

\begin{figure}[tbp]
  \begin{center}
\includegraphics[width=0.45\textwidth]{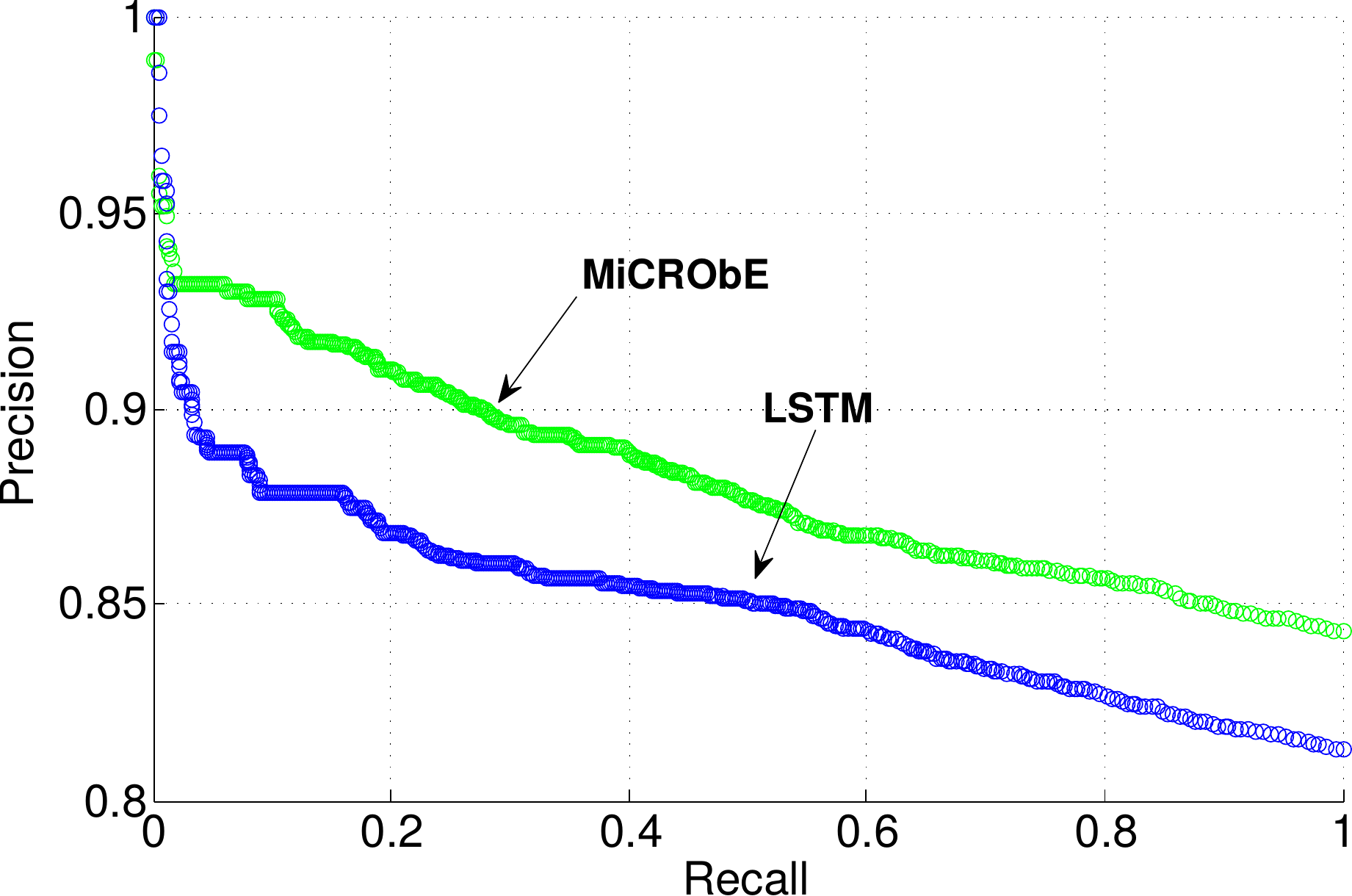}
\end{center}
\caption{The ROC for frame-level predictions for two models using human ratings for ground truth.}
\label{figure:roc}
\vspace{-1em}
\end{figure}

\begin{table*}[tbp]
\begin{center}
\begin{tabular}{|c|c|c|c|c|c|c|c|}
\hline
Features & Benchmark & \shortstack{\textbf{MaxCal} \\ ~} & \shortstack{\textbf{Logit} \\ \small{Hard Negs.}}  & \shortstack{\textbf{MiCRObE} (1 mix) \\ \small{Hard Negs}.} &  \shortstack{\textbf{MiCRObE} (5 mix) \\ \small{Hard Negs}.}  & \shortstack{\textbf{LSTM} \\ ~} \\
\hline
\textbf{IM} & \multirow{3}{*}{YT-12M} & 20.0\% & ~ & 36.2\%  & 36.6\% & 44.4\% \\
\cline{1-1}\cline{3-7}
\textbf{TM} &  & 28.0\% & ~ & 47.3\% & 47.3\% & 45.7\% \\
\cline{1-1}\cline{3-7}
\textbf{IM}$+$\textbf{TM} & & 29.0\% & 49.3\% & 50.1\%  & 49.5\% & 52.3\% \\
\hline
\textbf{IM} & \multirow{3}{*}{Sports-1M} &  28.2\% & 45.0\% & 46.5\%  & 47.2\% & 52.8\%\\
\cline{1-1}\cline{3-7}
\textbf{TM} &  &  38.6\% & 54.5\% & 55.4\% &  56.0\% & 58.8\% \\
\cline{1-1}\cline{3-7}
\textbf{IM}$+$\textbf{TM} & & 40.3\% & 54.7\% & 56.8\% & 57.0\% & 59.0\% \\
\hline
\end{tabular}
\end{center}
\caption{Hit@1 for the video level models against the ground truth.}
\label{tab:video_precision}
\vspace{-1em}
\end{table*}

We partition the training data using a $90/10$ split. This results in about $10.8$ million videos in the training partition and $1.2$M videos in the test partition. The ground-truth is only available at video-level. We train two kinds of models:

\textbf{Frame level models}: These models are trained to predict the label from a single frame. To provide robustness, a contagious idea is to feed MiCRObE the aggregated features over more than one frame. The features for each frame are obtained by averaging the features in a $\pm 2$ second window. For training the max-calibration model, we will use all the frames from each video in our training set and assume that every frame is associated with the video level ground-truth labels. For mining the collection of hard negatives and good positives for the second stage model, we randomly sample 10 frames from each video and mine the top $100,000$ scoring examples for each label from the resulting $108$ million frames (where the scoring is done using the maxcal model). At the inference time, we annotate each frame (sampled at 1fps) in the video using the trained MiCRObE cascade model.  The output of the LSTM model is evaluated at every frame, while passing the state forward.

Since we don't have frame level ground truth at such a large scale, we either (a) convert the frame level labels to the video level labels using the max-aggregation of the frame level probabilities and evaluate against the video-level ground truth (See Table \ref{tab:afp}), or (b) send a random sample of frames from a random sample of output labels to human raters (Figure~\ref{figure:roc}).

Note that the predictions of the underlying base models are entities which have some overlap with the target vocabulary. The precision numbers in the \textbf{Self} column are the accuracy of the base classifiers by themselves. For the combined model \textbf{IM+TM}, we take the maximum score of an entity coming from either of the models (Table~\ref{tab:afp}).

In order to prepare the data for human rating, we took a random set of $6,000$ videos which did not appear in the training set. For each video, we computed the output probabilities for all labels. For those labels which had an output probability of greater than $0.1$, we took all the frames which passed the thresholding, sorted the scores, and split the entire score range into $25$ equally sized buckets. From each bucket, we randomly sampled a frame and a score. For each model we evaluated, we randomly sampled $250$ labels (with $25$ frames each), and sent this set to human raters. The total number of labels from which we sampled these $250$ was $3,541$ for MiCRObE, and $1,568$ for the LSTM model. The resulting ROC is depicted in Figure~\ref{figure:roc}. We only considered frames for which there was a quorum (at least $2$ raters had to agree).

The MiCRObE method is well suited for frame-level classification due to
the fact that during the learning process it actively uses frame-level
information and mines hard examples. As such, it provides better performance than
the LSTM method on this task (Figure~\ref{figure:roc}).

\textbf{Video level models}: These models are trained to predict the labels directly from the aggregated features from the video.  The sparse features (available at each frame) are aggregated at the video level and the fusion models are directly trained to predict video-level labels from the (early) aggregated features. For this early feature aggregation, we collect feature specific statistics like mean, top-$k$ (for $k=1,2,3,4,5$) of each feature over the entire video. For example the label ``soccer'' from the TM model will expand to six different features \textit{TM:Soccer:Mean} (which is the average score of this feature over the entire video), \textit{TM:Soccer:1} (which is the highest score of this feature over the video), \textit{TM:Soccer:2} (which is the second highest score of this feature) and so on. The LSTM model remains unchanged from the frame-level prediction task. The video-level label is obtained by averaging the frame-level scores. The results are summarized in Table~\ref{tab:video_precision}.

On the Sports-1M benchmark, which is video-level, the LSTM method yields 59.0\%
hit@1. Karpathy~\etal~\cite{karpathy2014large} report 60.9\%, while
Tran~\etal~\cite{tran14corr} report 61.1\% using a single network which was
trained specifically for the task, starting from pixels. Similarly,
Ng~\etal~\cite{ng2015beyond} report 72.1\%. However, in order to obtain a
single prediction for the video, the video is passed through the network 240
times, which would not be possible in our scenario, since we are concerned
about both learning and inference speed.

 In terms of video classification, MiCRObE was adapted to use feature
 aggregation and it provides comparable performance to LSTM model (a hit@1
 within 2.8\% on YT-12M, and within 3\% on Sports-1M). The LSTM model, unlike
 MiCRObE, learns a representation of the input labels, and makes use of the
 sequential nature of the data.

Compared to previous work concentrating on large video classification, our
methods do not require training CNNs on the raw video data, which is desirable
when working with large numbers of videos. Our best performing single-pass
video-level model is within $2.1\%$ hit@1 of the best published model which
does not need multiple model evaluations per frame~\cite{tran14corr} (trained
directly from frames).

\vspace{-1em}
\section{Conclusion}
\vspace{-0.5em}
We studied the problem of efficient large scale video classification (12-million videos)
with a large label space (150,000 labels). We proposed to use image-based
classifiers which have been trained either on video thumbnails or on Flickr
images in order to represent the video frames, thereby avoiding a costly
pre-training step on video frames.  We demonstrate that we can organically
discover the correlated features for a label using the max calibration model.
This allows us to bypass the curse of dimensionality by providing a small set
of features for each label. We provided a novel technique for hard negative
mining using an underlying max-calibration model and use it to train a second
order mixture of experts model. MiCRObE can be used as a frame-level
classification method that does not require human-selected, frame-level ground
truth. This is crucial when attempting to classify into a large space of
labels. MiCRObE shows substantial improvements in precision of the learnt
fusion model against other simpler baselines like max-calibration and models
trained using random negatives and provides the highest level of performance at
the task of frame-level classification. We also show how to adapt this model
for video classification. Finally, we provide an LSTM based model that is
capable of the highest video-level performance on YT-12M. Performance
could further be improved by late-fusing outputs of the two algorithms.

{\small
\bibliographystyle{ieee}
\bibliography{egbib}
}

\end{document}